\documentclass{article}

    \PassOptionsToPackage{numbers, compress}{natbib}

    \usepackage[preprint]{neurips_data_2024}

\usepackage[utf8]{inputenc} 
\usepackage[T1]{fontenc}    
\usepackage[hidelinks]{hyperref}       
\usepackage{url}            
\usepackage{booktabs}       
\usepackage{amsfonts}       
\usepackage{nicefrac}       
\usepackage{microtype}      
\usepackage{xcolor}         
\usepackage{booktabs}
\usepackage{graphicx}
\usepackage{subfigure}

\usepackage{enumitem}
\usepackage{multirow}
\usepackage{algorithm}
\usepackage{algorithmic}
\usepackage{amsmath}
\usepackage{lineno}

\setitemize{label=\textbullet, leftmargin=*, nolistsep}

\title{A Benchmark for Maximum Cut: Towards Standardization of the Evaluation of Learned Heuristics for Combinatorial Optimization}

\author{%
  Ankur Nath \\
  Texas A\&M University \\
  College Station, TX, USA \\
  \texttt{anath@tamu.edu} \\
  \And
  Alan Kuhnle \\
  Texas A\&M University \\
  College Station, TX, USA \\
  \texttt{kuhnle@tamu.edu} \\
}

\begin{document}

\maketitle
\begin{abstract}
  Recently, there has been much work on the design of general heuristics for graph-based, combinatorial optimization problems 
  via the incorporation of Graph Neural Networks (GNNs) to learn distribution-specific solution structures.
  However, there is a lack of consistency in the evaluation of these heuristics, in terms of the baselines
  and instances chosen, which makes it difficult to assess the relative performance of the algorithms.
  In this paper, we propose an open-source benchmark suite \textbf{MaxCut-Bench} dedicated to the NP-hard Maximum Cut problem in both its weighted and unweighted variants, based on a careful selection of instances curated from diverse graph datasets. The suite offers a unified interface to various heuristics, both traditional and machine learning-based. Next, we use the benchmark in an
  attempt to systematically corroborate or reproduce the results of several, popular learning-based approaches,
  including S2V-DQN \citep{khalil2017learning}, ECO-DQN \citep{barrett2020exploratory}, among others, in terms of
  three dimensions: \textbf{objective value}, \textbf{generalization}, and \textbf{scalability}.
  Our empirical results show that several of the learned heuristics fail to outperform
  a naive greedy algorithm, and that only one of them consistently outperforms Tabu Search,
  a simple, general heuristic based upon local search. 
  Furthermore, we find that the performance of ECO-DQN remains the same
  or is improved if the GNN is replaced by a simple linear regression on a subset of the features that are related to Tabu Search.
  Code, data, and pretrained models are available at: \url{https://github.com/ankurnath/MaxCut-Bench}.


\end{abstract}


\section{Introduction}

The design of effective heuristics or approximation algorithms for NP-hard
combinatorial optimization (CO)  problems is a challenging task,
often requiring domain-specific knowledge alongside a rigorous process of empirical
refinement. Typically, the precise probability distribution of a particular set of
instances that are needed for a given application is complex or unknown, and may deviate
far from the set of worst-case instances that give rise to the computational complexity
of the problem at hand. For example, consider a shipping company that must solve a presumably
similar optimization problem each day for the routing of its delivery vehicles.
Consequently, there has been significant interest among researchers in automating
this demanding and tedious design process using machine learning to develop algorithms that exploit the
inherent structure of these distributions \citep{khalil2017learning,barrett2020exploratory,barrett2022learning,zhang2023let,tonshoff2022one}. Empirical evidence suggests that learned heuristics \citep{barrett2022learning,tonshoff2022one} can be competitive with state-of-the-art (SOTA) heuristics specifically tailored to individual problems.

However, along with the surge of automated, general heuristics for CO problems,
it is necessary to have a standardized way to evaluate these heuristics to determine what gains, if any,
are made over more traditional heuristics. The current status of the field is that each work
formulates a heuristic and then selects
its own baselines, problems, and instance distributions to evaluate
the heuristic -- inevitably, these are ones that showcase the new heuristic
in the best possible light. Often, a combination of
weak traditional heuristics and expensive exact algorithms are employed as baselines,
such as the Greedy algorithm (weak) or exact IP solvers, such as Gurobi \citep{gurobi} and CPLEX \citep{cplex2009v12}.
However, moderate-strength heuristics such as local search algorithms or even naive
stochastic greedy algorithms are typically omitted. 
More discussion and details can be found in Appendix \ref{appendix:Instance and Selection Bias}.

In this work, we focus on the Maximum Cut problem.
Many real-world applications \citep{perdomo2012finding,elsokkary2017financial,venturelli2019reverse}
can be reduced to MaxCut. Significant commercial and research efforts have been dedicated to developing MaxCut solvers, based on classical \citep{goto2019combinatorial} and quantum annealing \cite{leleu2019destabilization,tiunov2019annealing} approaches, as well as classical \citep{goemans1995improved,benlic2013breakout} and learned \citep{barrett2020exploratory,barrett2022learning,zhang2023let} algorithms. These efforts underscore the combination of intractability and broad applicability that motivates our focus on this problem. 
We carefully select distributions of instances for MaxCut -- then, we evaluate the performance of
several highly cited and recently proposed heuristics on these instances. 
We have the following research questions:

\begin{center}
  \textit{In this setting, can we reproduce or corroborate the performance of
    learned heuristics as compared to their traditional counterparts? Is there any absolute
    performance gain with learned heuristics compared to reasonably effective, traditional baselines? Finally, how well do algorithms trained with one distribution generalize to another? }
\end{center}


The Maximum Cut (MaxCut) problem is formally defined as follows. Given an undirected graph $G (V, E)$, where $V$ represents the set of vertices, $E$ denotes the set of edges and weights $w(u,v)$ on the edges $(u,v)\in E$, the goal is to find a subset of nodes $S \subseteq V$ that maximizes the objective function, $f(S)= \sum_{u \in S, v \in V\setminus S} w(u, v)$.

\textbf{Contributions.}
\begin{itemize}
\item We provide an open-source benchmark suite (\textbf{MaxCut-Bench}) for Maximum Cut solvers.
  The software currently supports several highly cited or recently proposed, learned heuristics such as S2V-DQN \cite{khalil2017learning}, ECO-DQN \citep{barrett2020exploratory}, LS-DQN \citep{yao2021reversible}, ANYCSP \citep{tonshoff2022one}, RUN-CSP \citep{toenshoff2021graph}, Gflow-CombOpt \citep{zhang2023let}, CAC \citep{leleu2021scaling}, and AHC \citep{leleu2019destabilization}. We reimplemented several learned local search algorithms using modern,
  more efficient graph learning packages, allowing them to scale up to larger instances.
  We propose a large and diverse collection of MaxCut instances with best known solution values, including
  instances that have been used in the past to benchmark SOTA traditional heuristics. 

    \item  Using \textbf{MaxCut-Bench}, we conduct an evaluation of learned and classical heuristics across multiple datasets.
      We show that Tabu Search \citep{glover1990tabu},
      a simple heuristic based upon local search, outperforms all the evaluated learned heuristics in
      terms of \textbf{objective value}, \textbf{scalability} and \textbf{generalization}, except for ANYCSP. Moreover, a naive reversible
      greedy algorithm outperforms several of the learned heuristics and is competitive with the popular
      S2V-DQN algorithm. 
      Due to the wide-ranging solid performance of Tabu Search across all instances, we recommend that it be
      included as a baseline for any future heuristic claimed to solve MaxCut.
      ANYCSP, a global search heuristic, is the best performing of the evaluated heuristics, 
      indicating that a reinforcement learning approach to general constraint satisfaction for CO problems
      is a promising direction for future research.

    \item Using \textbf{MaxCut-Bench},
      we show that while combining local search algorithms with deep learning can improve their performance, the good performance of a highly-cited heuristic, ECO-DQN,
      can be achieved by simply selecting a subset of its features (which are related to Tabu Search)
      and replacing the GNN with a linear regression model.
      This result suggests that care should be taken not to conflate the performance added by the GNN with the power
      of an underlying classical heuristic, such as local search or Tabu Search. 
    
\end{itemize}
We believe that this benchmark will enable research community to standardize further research endeavors in machine learning for solving CO problems .

\textbf{Organization.} The rest of this paper is organized as follows. In Section \ref{sec:rw}, we discuss relevant related work. We present the the MaxCut-Bench instance distributions and algorithms in Section \ref{sec:benchmark}. In Section \ref{sec:eval}, we use the benchmark to answer the motivating questions above. Finally, in Section \ref{sec:conclusion}, we conclude the paper. 


\section{Related Work} \label{sec:rw}
Alongside the growing interest in GNNs for tackling CO problems,
there have been a number of recent works revisiting the effectiveness
versus traditional heuristics. 
\citet{angelini2023modern} demonstrated that a simple greedy algorithm running almost in linear time can find solutions of
better quality than a physics-inspired unsupervised GNN \citep{schuetz2022combinatorial} for the Maximum Independent Set (MIS) problem. Similarly, for MIS, \citet{bother2022s} demonstrated that the performance of the popular guided tree search algorithm \citep{li2018combinatorial} is not reproducible, and the GNN used in the tree search does not play any meaningful role. Rather, the various classical algorithms are the reason for good performance, especially on hard instances. This is analogous to our result for the ECO-DQN algorithm (Section \ref{section:ablation}), where we show that the GNN does not appear to play a meaningful role in the algorithm, and instead, the good performance of ECO-DQN for MaxCut may come from its Tabu Search-related features.

For Boolean Satisfiability, RLSAT \citep{yolcu2019learning} was shown to match the performance of the traditional local search algorithm WalkSAT \citep{selman1993local}. However, \citet{tonshoff2022one} demonstrated that WalkSAT easily outperforms RLSAT on hard instances. In the context of the Traveling Salesman problem (TSP), \citet{joshi2020learning} noted that GNNs can achieve promising results for relatively small instances, typically up to a few hundred cities. However, for instances involving millions of cities, the classical Lin-Kernighan-Helsgaun algorithm \citep{helsgaun2000effective,taillard2019popmusic} consistently finds solutions close to optimality. More recently, \citet{liu2023good} demonstrated that learned heuristics still lag behind traditional solvers in effectively solving TSP. Regarding the MaxCut problem, \citet{yao2019experimental} showed that a straightforward local search algorithm, Extremal Optimization \citep{boettcher2001extremal}, can outperform a specific GNN \citep{chen2017supervised} across various configurations of dense and sparse random regular graphs.
In contrast to our work, their study focused solely on random regular graphs and did not include comparisons with existing SOTA learned heuristics from the literature.

\section{The MaxCut-Bench Benchmark} \label{sec:benchmark}
In this section, we introduce the datasets, distributions, and algorithms included in \textbf{MaxCut-Bench}.
\subsection{Benchmark Datasets}
We have curated a diverse collection of instances from both real-world datasets and random graph distributions to ensure a comprehensive and rigorous evaluation. To ensure a thorough comparison, we use commonly employed random graph models for instance generation, including \citet{erdHos1960evolution} (ER), \citet{albert2002statistical} (BA), \citet{holme2002growing} (HK), and \citet{watts1998collective} (WS).
For harder distributions, we use the following:
Sherrington-Kirkpatrick (SK) spin glass \citep{sherrington1975solvable}, dense unweighted instances at their phase transitions \citep{coppersmith2004random}, Physics \citep{festa2002randomized}, and the Stanford GSet dataset \citep{ye2003gset}.
These harder instances have been employed for evaluating the performance of several SOTA traditional heuristics \citep{benlic2013breakout,leleu2019destabilization,leleu2021scaling,hamerly2019experimental}. 
Additional details regarding all mentioned datasets and the hyperparameters used for graph generation can be found in Appendix \ref{appendix:datasets}. These datasets consist of graphs ranging in size from $70$ to $2000$ vertices.

\subsection{Benchmark Algorithms}

In this section, we give an overview of the different algorithms available in the benchmark. For more discussion of these algorithms, please refer to Appendix \ref {appendix:algorithms}. For our study, we consider both traditional and learning-based models. They can be organized into four categories.

\textbf{Traditional Heuristics.}
These algorithms start with a candidate solution and then iteratively move to a neighboring solution by adding or removing a vertex. We provide implementations of the following algorithms: \textbf{Forward Greedy (FG)}, the traditional simple greedy heuristic that starts with an empty solution adds the largest gain to the candidate solution; \textbf{Reversible Greedy (RG)}, a variant of forward greedy that starts with a random solution and is also allowed to remove elements from the solution if gain in objective value can be obtained from such removal; 
\textbf{Tabu Search (TS)} \citep{glover1990tabu}, a modification of RG that avoids solutions previously seen to promote a more diverse search;
and \textbf{Extremal optimization (EO)}\cite{boettcher2001extremal}, a stochastic local search algorithm. 

\textbf{Quantum Annealing (QA) Heuristics.}
These algorithms are designed to find the global minimum of an objective function within a set of potential solutions and are used mainly for problems where the search space is discrete (CO problems) with many local minima, such as finding the ground state of a spin glass or solving the traveling salesman problem. Since the MaxCut problem is equivalent to minimizing the Hamiltonian of a spin glass model \cite{barahona1988application}, our benchmark supports two 
annealing algorithms, \textbf{Amplitude Heterogeneity Correction (AHA)}  \citep{leleu2019destabilization}, and
\textbf{Chaotic Amplitude Control (CAC)}  \citep{leleu2021scaling}. We remark that none of the heuristics evaluated
in this work exceed these QA heuristics in objective value; therefore,
in the following we normalize by the best solution value found by one of these algorithms. 

\textbf{GNN-based heuristics.} 
\citet{khalil2017learning} introduced \textbf{S2V-DQN}, which has become a highly influential work with
thousands of citations. It may be thought as the forward greedy algorithm, except a GNN is used to select the next element to include. We reimplement this algorithm along with reversible extensions of S2V-DQN: 
\textbf{ECO-DQN} \cite{barrett2020exploratory}  and
\textbf{LS-DQN} \cite{yao2021reversible}. 

We include \textbf{Gflow-CombOpt} \cite{zhang2023let},
a recently published spotlight work at NeurIPS 2023.
This algorithm formulates CO problems as sequential decision-making sampling problems and designs efficient conditional Generative Flow Networks \cite{bengio2021flow} to sample from the solution space and generate diverse solution candidates. 

Finally, as MaxCut can easily be reduced to a constraint satisfaction problem, we include \textbf{RUN-CSP} \citep{toenshoff2021graph}
and \textbf{ANYCSP} \citep{tonshoff2022one}. These algorithms allows transitions between any two solutions in a single step.

\section{Evaluation} \label{sec:eval}

In this section, we use the benchmark to investigate the {\textbf{objective value}}, {\textbf{scalability}} and {\textbf{generalization}} of algorithms by answering the following specific questions:

\begin{itemize}
    \item For {\textbf{objective value}}, does combining deep learning with simple heuristics improve performance, or can the simple heuristics, with a much simpler model, do the heavy lifting (Section \ref{section:ablation})? Secondly, can the learned heuristics outperform simple, general heuristics like Reversible Greedy and Tabu Search (Section \ref{section:bias})?

    \item For {\textbf{generalization}}, how well do these learning-based algorithms generalize beyond their training data on out-of-distribution instances (Section \ref{generalization})?

     \item For {\textbf{scalability}}, how efficient are these algorithms in terms of time and space on large hard instances? Is the trade-off between complexity and performance reasonable (Section \ref{section:scalability})?
\end{itemize}

All learned algorithms in our benchmark are implemented using PyTorch \citep{paszke2019pytorch}. All experiments were conducted on a Linux server with a GPU (NVIDIA RTX A6000) and CPU (AMD EPYC 7713), using PyTorch 2.3.0, CIM \cite{Chen_cim-optimizer_a_simulator_2022}, DGL 2.2.1 \cite{wang2019deep}, and Python 3.11.9. We clearly state the changes necessary to update previous implementations and ensure that our reimplementations align with the results published in the literature. We provide all the details of these changes in Appendix \ref{appendix:reproducibility}.

\textbf{Evaluation settings.}
All training is performed on randomly $4000$ generated graphs and the validation is performed on a fixed set of $50$ held-out graphs from the same distribution.
For synthetic datasets, testing is performed on $100$ instances drawn from the same distribution; or upon the test instances provided in the original resource (see details in Appendix \ref{appendix:datasets}).
We evaluate the algorithms using the average approximation ratio as a performance metric. Given an instance, the approximation ratio for an algorithm is computed by normalizing the objective value of the best-known solution for the instance. As S2V-DQN and Forward Greedy are deterministic at test time, we use a single optimization episode for each graph. For all other algorithms, we run each algorithm for $50$ randomly initialized episodes with $2|V|$ number of search steps per episode and select the best outcome from these runs following the experimental setup of \citet{barrett2020exploratory} and \citet{yao2021reversible}. Experimentally, we have found that the performance of all algorithms saturates within this number of search steps (we refer the reader to the Appendix \ref{appendix:evaluation_settings} for more details).

\subsection{Does deep learning really improve the performance of a traditional heuristic?} \label{section:ablation}
In this subsection, we address the question of whether the deep learning heuristics
that combine a deep neural network with a traditional heuristic actually improve
over the original heuristic. 

\textbf{S2V and LS-DQN.}
Specifically, the concept of S2V-DQN is exactly the Forward Greedy algorithm guided by a GNN.
Similar to Forward Greedy, S2V-DQN begins with an empty solution and incrementally add a vertex to the solution until no greedy actions remain. The key distinction lies in how they select vertices: while Forward Greedy opts for the vertex that maximizes objective value increase, S2V-DQN delegates this decision to its GNN. Following in the footsteps of S2V-DQN, LS-DQN seeks to enhance
vanilla Reversible Greedy. However, unlike vanilla RG, which halts when no greedy actions are left, LS-DQN incorporates the GNN to determine its termination criteria. Therefore, we seek to reproduce the results of
\citet{khalil2017learning,yao2021reversible}, which show
improvement over the respective traditional heuristic on a more limited selection of instances.

\textbf{ECO-DQN.} While S2V-DQN and LS-DQN use the current solution as the state-space representation for the reinforcement learning (RL) agent, ECO-DQN employs seven handcrafted features per vertex for its state space. Upon closer examination, we find that two of these features are closely related to the traditional heuristic, Tabu Search: \textbf{1) Marginal Gain:} the change in the objective value when a vertex is added to or removed from the solution set (we refer this action as a flip) \textbf{2) Time Since Flip:} steps since the vertex has been flipped to prevent short looping trajectories. To understand the connection between TS and ECO-DQN,
we delve into how TS functions. Similar, to ECO-DQN, TS starts with a random solution and iteratively explores the best neighboring solutions. To prevent revisiting local minima and explore the search space more effectively, TS employs a parameter known as tabu tenure, which acts as a short-term memory. This parameter restricts the repetition of the same actions for tabu tenure steps, preventing the algorithm from revisiting recent actions. Therefore, ECO-DQN uses a superset of the Tabu Search features to learn its agent. 

To better understand the performance of ECO-DQN, we compare with an ablated version, which we call SoftTabu,
which replaces the GNN with linear regression and omitting the features not related to TS, that is, all
but the two features described above.

\begin{figure}[h]
    \centering
    \includegraphics[width=\linewidth]{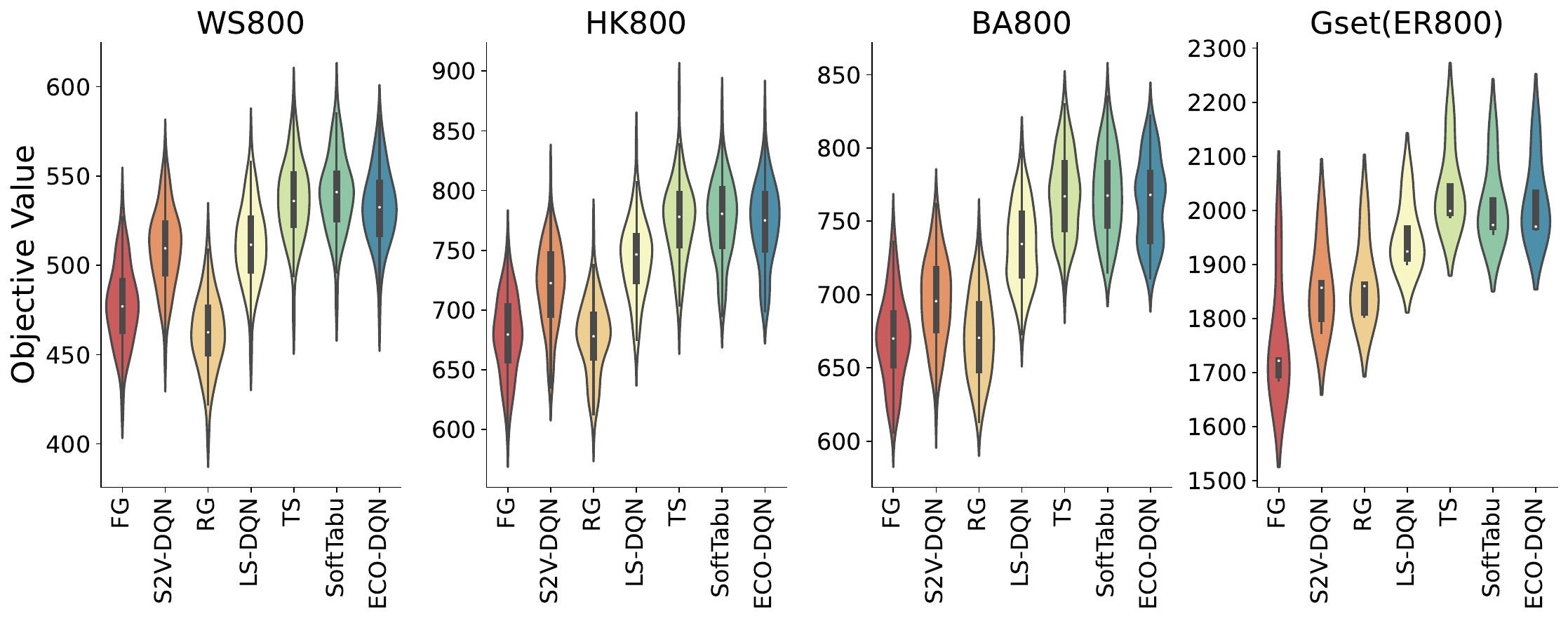}
    \caption{Violin plots of objective values of the learned and their classical counterparts on a selection of weighted instances.}
    \label{fig:violin-cdl}
\end{figure}

\begin{table}[h]
\centering
\caption{Performance comparison of learned heuristics with their simple counterparts (best in \textbf{bold}): The first and second halves of the table show results for unweighted and weighted instances, respectively.}
\label{tab:abalation}
\resizebox{\textwidth}{!}{%
\begin{tabular}{@{}cccccccccccc@{}}
\toprule
                    &         &  & \multicolumn{2}{c}{S2V-DQN} &  & \multicolumn{2}{c}{LS-DQN} &  & \multicolumn{3}{c}{ECO-DQN} \\ \cmidrule(lr){4-5} \cmidrule(lr){7-8} \cmidrule(l){10-12} 
Graph               & Nodes   &  & FG       & FG+GNN      &  & RG       & RG+GNN    &  & TS    & SoftTabu & ECO+GNN   \\ \midrule
GSet (ER)& 800     &  & $0.970^{\pm0.003}$ & $0.970^{\pm0.001}$&  & $0.985^{\pm0.001}$&   $0.992^{\pm0.001}$&  &  $ \mathbf{0.999^{\pm0.001} } $ &   $0.996^{\pm0.001}$& $0.997^{\pm0.001}$\\
GSet (Skew)         & 800     &  & $0.960^{\pm0.002}$& $0.980^{\pm0.001}$&  & $0.963^{\pm0.002}$&   $0.984^{\pm0.002}$&  & $ \mathbf{0.992^{\pm0.001}} $&    $0.991^{\pm0.001}$&    $0.990^{\pm0.001}$\\
BA                  & 800     &  &         $0.938^{\pm0.007}$&        $0.951^{\pm0.006}$&  &        $0.952^{\pm0.003}$&         $0.983^{\pm0.003}$&  &      $0.992^{\pm0.002}$&       $\mathbf{0.993^{\pm0.002}} $&         $0.991^{\pm0.002}$\\
WS                  & 800     &  &         $0.944^{\pm0.006}$&         $0.973^{\pm0.005}$&  &        $0.927^{\pm0.003}$&              $0.971^{\pm0.003}$&  &       $0.990^{\pm0.002}$&         $0.988^{\pm0.002}$&           $ \mathbf{ 0.992^{\pm0.002}}   $\\
HK                  & 800     &  &         $0.939^{\pm0.007}$&         $0.966^{\pm0.005}$&  &        $0.951^{\pm0.003}$&         $0.984^{\pm0.003}$&  &       $ \mathbf{0.992^{\pm0.002}}  $ &         $ \mathbf{0.992^{\pm0.002}}  $  &           $ \mathbf{0.992^{\pm0.002}}  $   \\
Phase Transition    & 100-200 &  & $0.982^{\pm0.005}$&   $0.985^{\pm0.006}$&  & $0.996^{\pm0.002}$&  $0.998^{\pm0.002}$&  &     $ \mathbf{ 1.000^{\pm0.000}}   $  &    $ \mathbf{1.000^{\pm0.000}}  $&     $ \mathbf{1.000^{\pm0.000}} $\\ \bottomrule \midrule 
GSet (ER)           & 800     &  &   $0.856^{\pm0.026}$&     $0.906^{\pm0.017}$&  &    $0.911^{\pm0.012}$&   $0.953^{\pm0.007}$&  &  $\mathbf{0.995^{\pm0.003}}$&    $0.981^{\pm0.002}$&     $0.984^{\pm0.003}$\\
GSet (Skew)         & 800     &  &    $0.863^{\pm0.030}$&    $0.890^{\pm0.027}$&  &     $0.871^{\pm0.013}$&     $0.942^{\pm0.007}$&  &   $ \mathbf{0.980^{\pm0.005}}   $&     $0.975^{\pm0.005}$&      $0.966^{\pm0.009}$\\
GSet (Torodial)     & 800     &  &   $0.838^{\pm0.003}$&      $0.960^{\pm0.010}$&  &     $0.793^{\pm0.010}$&     $0.965^{\pm0.007}$&  &   $0.989^{\pm0.003}$ &    $0.993^{\pm0.004}$ &    $ \mathbf{0.994^{\pm0.002}}  $\\ 
BA                  & 800     &  &              $0.853^{\pm0.018}$&              $0.886^{\pm0.019}$&  &             $0.855^{\pm0.012}$&              $0.934^{\pm0.011}$&  &       $ 0.978^{\pm0.007} $&         $ \mathbf{0.979^{\pm0.008}} $&           $0.972^{\pm0.009}$\\
WS                  & 800     &  &        $0.861^{\pm0.014}$&        $0.919^{\pm0.012}$&  &        $0.833^{\pm0.010}$&              $0.923^{\pm0.007}$&  &       $0.967^{\pm0.006}$&         $ \mathbf{0.973^{\pm0.006}}  $&           $0.961^{\pm0.008}$\\
HK                  & 800     &  &              $0.857^{\pm0.019}$&              $0.908^{\pm0.017}$&  &             $0.855^{\pm0.012}$&              $0.937^{\pm0.009}$&  &       $ 0.977^{\pm0.007}$&         $\mathbf{0.978^{\pm0.008}}$&           $0.974^{\pm0.009}$\\
Barrett et al. (ER) & 200     &  &  $0.866^{\pm0.038}$&   $0.951^{\pm0.024}$&  &  $0.954^{\pm0.014}$&   $0.987^{\pm0.010}$&  &   $ \mathbf{1.000^{\pm0.001}} $&   $ \mathbf{1.000^{\pm0.001}} $ &   $ \mathbf{1.000^{\pm0.001}} $\\
Barrett et al. (BA) & 200     &  &  $0.849^{\pm0.054}$&   $0.937^{\pm0.043}$&  &   $0.903^{\pm0.039}$&   $0.977^{\pm0.032}$&  &  $ \mathbf{0.984^{\pm0.032}}  $ &   $\mathbf{0.984^{\pm0.032}} $ &   $0.983^{\pm0.033}$\\
SK spin-glass       & 70-100  &  &  $0.865^{\pm0.057}$&   $0.939^{\pm0.049}$&  &    $0.994^{\pm0.010}$&   $0.999^{\pm0.003}$&  &  $ \mathbf{1.000^{\pm0.000}} $&   $ \mathbf{1.000^{\pm0.000}} $ &   $ \mathbf{1.000^{\pm0.000}} $\\
Physics (Regular)   & 125     &  &   $0.779^{\pm0.049}$&   $0.962^{\pm0.023}$ &  &    $0.872^{\pm0.022}$ &   $0.991^{\pm0.010}$&  &  $ \mathbf{1.000^{\pm0.000}} $ &    $ \mathbf{1.000^{\pm0.000}} $ &    $ \mathbf{1.000^{\pm0.000}} $ \\

\bottomrule

\end{tabular}%
}
\end{table}

\textbf{Impact of the Findings.} We evaluate these learned heuristics in various configurations on a wide range of graph distributions, with results shown in Table \ref{tab:abalation} and Figure \ref{fig:violin-cdl}.
In summary, the GNN is shown to offer improvement when combined with the simpler classical heuristics Forward and Reversible Greedy;
however, ECO-DQN fails to consistently improve on TS. In fact, the very simple heuristic TS outperforms all of the algorithms evaluated in this subsection.
Next, we delve deeper to investigate the causes of some of the unexpected outcomes.  

\begin{figure}[h]
    
     \centering
     \subfigure { 
        \includegraphics[width=\linewidth]{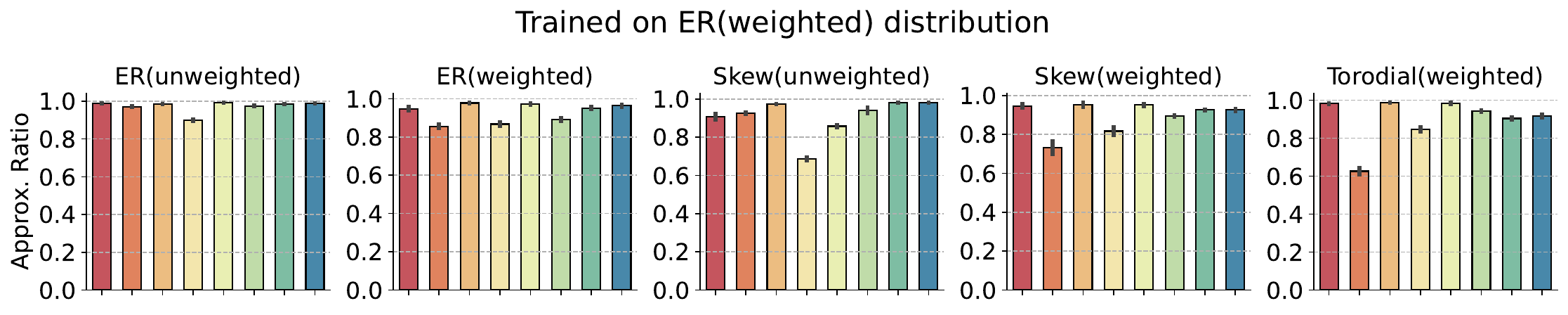}
    }
    \subfigure { 
    \includegraphics[width=\linewidth]{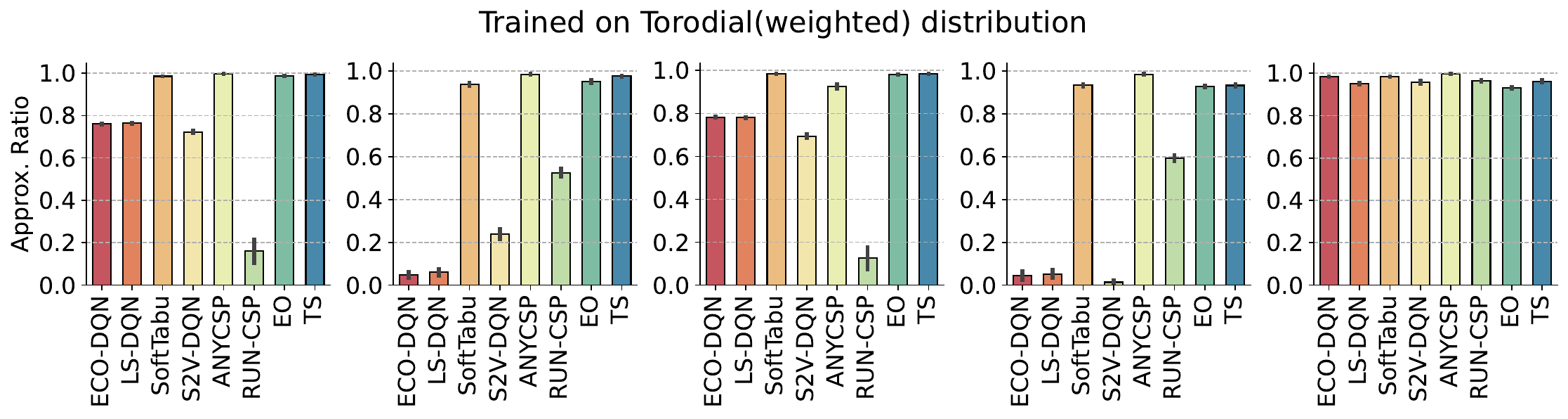}
  }
     
     \caption{Generalisation of agents to unseen graph sizes and structures.}
\label{fig:generalization}

\end{figure}

\begin{itemize}
\item \textbf{S2V-DQN.} From Table \ref{tab:abalation}, it is evident that S2V-DQN (FG+GNN) significantly improves upon the Forward Greedy approach for the MaxCut problem, corroborating the empirical evaluation presented in the original work.
  This improvement is impressive, showing the promise of using GNNs to improve classical heuristics.
  However, we observe that TS, a simple and general heuristic with a single parameter, consistently outperforms S2V-DQN.
  In fact, even RG outperforms S2V-DQN on some distributions, such as ER (weighted and unweighted),
  BA (unweighted), Phase Transition networks
  among others, as shown in Table \ref{tab:abalation}. \textcolor{black}{ This demonstrates how stochasticity is a powerful attribute when combined with local search. \citet{khalil2017learning} showed that the simple greedy approach, which starts with an empty solution and greedily moves the node that results in the largest improvement in cut weight, was the second-best competitor to S2V-DQN but did not compare it with RG.}


\item \textbf{ECO-DQN.} From Table \ref{tab:abalation},  we observe that TS and SoftTabu often match or outperform ECO-DQN (ECO+GNN) in terms of performance. Similar to ECO-DQN, SoftTabu learns to strike a balance between exploration and exploitation. To ensure a fair comparison, we also used the publicly available datasets used to evaluate ECO-DQN in the original paper
  of \citet{barrett2020exploratory}, which consists of ER and BA graphs.

  In addition, we conducted all empirical experiments presented in ECO-DQN and compared them with TS and SoftTabu (additional results can be found in Appendix \ref{appendix:eco_dqn}). We found no significant difference in performance between ECO-DQN and its simpler counterparts.
  We conclude that the major contributing factors to solution quality can be achieved by the features related to TS with a simple model, and ECO-DQN does not gain any significant performance boost from being guided by a deep learning model.
  This underscores the importance of comparing an algorithm on a range of instances, as very small, synthetic generated datasets may lack the capability to effectively differentiate algorithm performance. In fact, adding the GNN adversely
  affects the generalization performance of ECO-DQN (as discussed in Section \ref{generalization}, Figure \ref{fig:generalization}). 

\item \textbf{LS-DQN.}  Our results support that
  LS-DQN enhances the performance of RG, as reported in \citet{yao2021reversible}.
  However, our analysis contradicts the claim of \citet{yao2021reversible} that LS-DQN matches the performance of ECO-DQN.
  This may be explained by observing that \citet{yao2021reversible} compared the performance of ECO-DQN and LS-DQN
  only on complete graphs, which underscores the importance of comparing algorithms across a range of instances.
\end{itemize}

\textbf{Final Considerations.}
In summary, adding a GNN to a traditional heuristic showed an improvement in the cases of Forward Greedy and Reversible
Greedy. However, the authors of ECO-DQN added enough features to the algorithm to enable learning something
similar to TS, and the GNN does not add any additional improvement over TS and may in fact hurt the performance.


\subsection{Have deep learning heuristics obtained any absolute improvement over the best traditional heuristic?} \label{section:bias} 
In this section, we evaluate the algorithms to assess whether any of the
deep learning heuristics can achieve the SOTA
objective value on the instances in the MaxCut benchmark.
While it may be too early to expect learned heuristics to beat SOTA heuristics tailored for specific problems, the question of how learned heuristics fare against general heuristics remains unanswered. 
One of the main reasons for this is the variety of baselines and instances used in existing work (more details can be found in Appendix \ref{appendix:Instance and Selection Bias}). 

\begin{figure}[h]
    \centering
    \includegraphics[width=\linewidth]{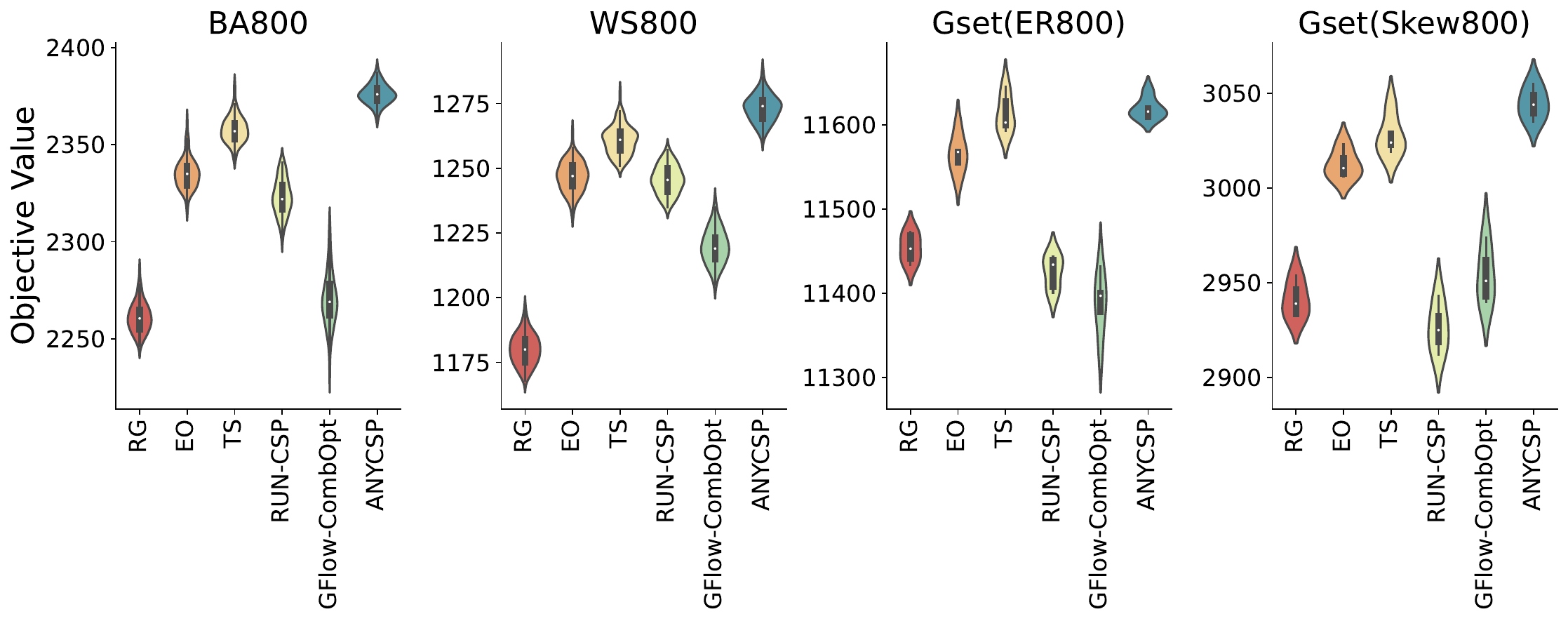}
    \caption{Violin plots of objective values of the learned and classical heuristics on a selection of unweighted instances. }
    \label{fig:violin}
\end{figure}

\begin{table}[]
\centering
\caption{Average Approximation ratios of classical and learned heuristics (best in \textbf{bold});``---" denotes no
reasonable result is achieved by the corresponding algorithm and the first and second halves of the table show results for unweighted and weighted instances, respectively.}
\label{tab:bias}
\resizebox{\textwidth}{!}{%
\begin{tabular}{@{}cccccccccc@{}}
\toprule
                    &   && \multicolumn{3}{c}{Classical Heuristics} &  & \multicolumn{3}{c}{Learned Heuristics} \\ \cmidrule(lr){3-6} \cmidrule(l){7-10} 
Graph               &   Nodes && RG       & EO         & TS          &  & RUN-CSP  & GFlow-CombOpt  & ANYCSP \\ \midrule
GSet (ER)           &  800 && $0.985^{\pm0.001}$& $0.995^{\pm0.001}$& $\mathbf{0.999^{\pm0.001}}$&  & $0.982^{\pm0.001}$& $0.979^{\pm0.003}$& $\mathbf{0.999^{\pm0.000}}$\\
GSet (Skew)         &  800  && $0.963^{\pm0.002}$& $0.987^{\pm0.000}$& $0.992^{\pm0.001}$&  & $0.958^{\pm0.003}$& $0.967^{\pm0.003}$& $\mathbf{0.997^{\pm0.001}}$\\
BA                  &  800 && $0.952^{\pm0.003}$& $0.983^{\pm0.003}$& $0.992^{\pm0.002}$&  & $0.978^{\pm0.003}$& $0.955^{\pm0.006}$& $\mathbf{1.000^{\pm0.000}}$\\
WS                  &  800 && $0.927^{\pm0.003}$& $0.979^{\pm0.003}$& $0.990^{\pm0.002}$&  & $0.978^{\pm0.003}$& $0.958^{\pm0.004}$& $\mathbf{0.999^{\pm0.000}}$\\
HK                  &  800 && $0.951^{\pm0.003}$& $0.983^{\pm0.003}$& $0.992^{\pm0.002}$&  & $0.977^{\pm0.003}$&  $0.930^{\pm0.015}$& $\mathbf{1.000^{\pm0.000}}$\\
Phase Transition    &  100-200 && $0.996^{\pm0.002}$& $0.999^{\pm0.001}$& $\mathbf{1.000^{\pm0.000}}$&  & $0.994^{\pm0.002}$& $0.993^{\pm0.004}$& $\mathbf{1.000^{\pm0.000}}$\\ \bottomrule \midrule
GSet (ER)           &  800 && $0.911^{\pm0.012}$&  $0.973^{\pm0.005}$& $0.995^{\pm0.003}$&  & $0.909^{\pm0.007}$& ---& $\mathbf{0.998^{\pm0.002}}$\\
GSet (Skew)         &  800 && $0.871^{\pm0.013}$& $0.954^{\pm0.007}$& $0.980^{\pm0.005}$&  & $0.925^{\pm0.013}$& ---& $\mathbf{0.995^{\pm0.005}}$\\
GSet (Torodial)     &  800 && $0.793^{\pm0.010}$& $0.948^{\pm0.010}$& $0.989^{\pm0.003}$&  &  $0.976^{\pm0.002}$& ---& $\mathbf{0.999^{\pm0.002}}$\\
BA                  &  800 && $0.855^{\pm0.012}$& $0.949^{\pm0.008}$& $0.978^{\pm0.007}$&  & $0.931^{\pm0.009}$& ---& $\mathbf{1.000^{\pm0.000}}$\\
WS                  &  800 && $0.833^{\pm0.010}$& $0.943^{\pm0.007}$& $0.967^{\pm0.006}$&  & $0.950^{\pm0.007}$& ---& $\mathbf{0.999^{\pm0.000}}$\\
HK                  &  800 && $0.855^{\pm0.012}$& $0.949^{\pm0.009}$& $0.977^{\pm0.007}$&  & $0.937^{\pm0.009}$& ---& $\mathbf{1.000^{\pm0.000}}$\\
Barrett et al. (ER) &  200 && $0.954^{\pm0.014}$& $0.989^{\pm0.008}$& $\mathbf{1.000^{\pm0.001}}$&  & $0.940^{\pm0.015}$& ---& $\mathbf{1.000^{\pm0.000}}$\\
Barrett et al. (BA) &  200 && $0.903^{\pm0.039}$& $0.969^{\pm0.035}$& $0.984^{\pm0.032}$&  & $0.958^{\pm0.035}$& ---& $\mathbf{0.986^{\pm0.032}}$\\
SK spin-glass       &  70-100 && $0.994^{\pm0.010}$& $0.995^{\pm0.006}$& $\mathbf{1.000^{\pm0.000}}$&  & $0.962^{\pm0.019}$& ---& $\mathbf{1.000^{\pm0.001}}$\\
Physics (Regular)   &  125 && $0.872^{\pm0.022}$ & $0.986^{\pm0.011}$& $\mathbf{1.000^{\pm0.000}}$&  & $0.989^{\pm0.009}$& ---& $\mathbf{1.000^{\pm0.000}}$\\ \bottomrule
\end{tabular}%
}
\end{table}

\textbf{Impact of findings.} We compare GFlow-CombOpt, RUN-CSP, and ANYCSP with the traditional heuristics
RG, EO, and TS. 
We observe that deep learning-based solutions can often be outperformed, especially on larger instances.
Next, we discuss the insights that our empirical observations can provide for each algorithm.

\begin{itemize}

\item \textbf{GFlow-CombOpt.} We restrict our empirical evaluation of GFlow-CombOpt to unweighted instances because \citet{zhang2023let} only evaluated GFlow-CombOpt for MaxCut on unweighted BA graphs, following \citet{sun2022annealed},
  and our attempt to adapt GFlow-CombOpt to handle weighted graphs (details in Appendix \ref{appendix:reproducibility}) performed poorly and did not surpass the Forward Greedy baseline.
  From Table \ref{tab:bias} and Figure \ref{fig:violin}, TS and EO consistently outperform GFlow-CombOpt.
  Moreover, surprisingly,  RG exceeds the performance of GFlow-CombOpt on four of the six distributions
  tested, including unweighted BA graphs.
  In our opinion, this example showcases the need for more standardization of evaluation,
  as GFlow-Combopt
  is outperformed more often than not by a naive greedy heuristic. 

\item \textbf{RUN-CSP.}  From Table \ref{tab:bias}, we observe that TS outperforms RUN-CSP on all distributions, EO outperforms RUN-CSP on most distributions, while RUN-CSP outperforms RG in most, but not all, distributions. Although \citet{toenshoff2021graph} included hard unweighted instances from the Gset dataset, they showed RUN-CSP obtains superior performance to a
  Semidefinite Programming (SDP) solver based on dual scaling \citep{choi2000solving}. 
  SDP-based approaches typically underperform in practice for MaxCut \citep{khalil2017learning, yao2021reversible}, even with extended computation times. Although these approaches have the highest theoretical approximation ratio, they are easy to beat in practice. Again, this underscores the need for carefully chosen baselines to demonstrate improvement.


\item    \textbf{ANYCSP.}  We analyze ANYCSP in its default configuration and observe that it consistently finds near-optimal solutions (that is, solutions near the value found by the Quantum Annealing algorithms), and superior in value to TS and the other traditional local search algorithms.
      These promising results demonstrate that ANYCSP can effectively learn the solution structure across diverse graph distributions and remains robust across multiple distributions .

\end{itemize}
\textbf{Final considerations.} The empirical findings suggest that benchmarking against weak heuristics on a very particular set of instances may establish a low standard, potentially leading to a misleading sense of achievement. Hence, selecting appropriate baselines is crucial to accurately assess the effectiveness of learned heuristics. At a minimum, we recommend that any work should at least include RG, and ideally TS, as a baseline -- failure to consistently beat the naive RG baseline suggests that the algorithm may need further development. 

\subsection{Generalization: Do learned heuristics generalizes well on unseen distributions?   } \label{generalization}
The ability of learned heuristics to perform well on a wide range of distributions, even if these distributions are not represented during training, is a highly desirable characteristic for practical CO problems \citep{cappart2023combinatorial}.
Many works include experiments to assess how well an the heuristic generalizes. Often, this is in the form of training
on smaller instances and generalizing to test on larger ones, although cross-distribution performance is also frequently
assessed.

Several learned heuristics such as ECO-DQN, LS-DQN, and S2V-DQN are claimed to exhibited promising performance across a diverse range of graph structures, including those not present in their training data.
In this section, we evaluate the generalization performance using \textbf{MaxCut-Bench}.

\textbf{Impact of Findings.} From Figure \ref{fig:generalization}, we notice that there can be a substantial decline in performance when the learned heuristics are tested on graph distributions other than train distributions, with the notable exception of ANYCSP. 
In particular, observe that when trained on the torodial distribution (second row), the test performance of ECO-DQN, LS-DQN, S2V-DQN, and RUN-CSP may fall below 25\% on several distributions. 
This outcome may be anticipated. Intuitively, we would expect a network trained on instances of a particular structure to adapt toward this class of instances and perform poorer for different structures.
We observe that TS, SoftTabu and EO seem to be generalize well across wider distributions. Both TS and EO has a single parameter, which we optimized for the training distribution to assess its generalization performance. 

These results raise the possibility that the generalization of learned heuristics from learning over small and easy instances to testing on larger and more complicated ones  may not be as robust as the literature \citep{khalil2017learning,barrett2020exploratory,barrett2022learning,yao2021reversible} suggests. This feature is often touted as an amelioration of the expensive training process required for the learned heuristics. We provide additional results on our generalization experiments in \ref{appendix:generalization}.


\subsection{Efficiency and scalability analysis} \label{section:scalability}
In this section, we analyze the efficiency and scalability of learned heuristics over ER graphs of size $|V|=800$ from Gset dataset. For time efficiency, we evaluate the efficiency of the algorithms by measuring the wall-clock time. For scalability, we evaluate average GPU and CPU usage per second of these learned algorithms. 

\begin{figure}[h]

     \centering

    \includegraphics[width=0.95\linewidth]{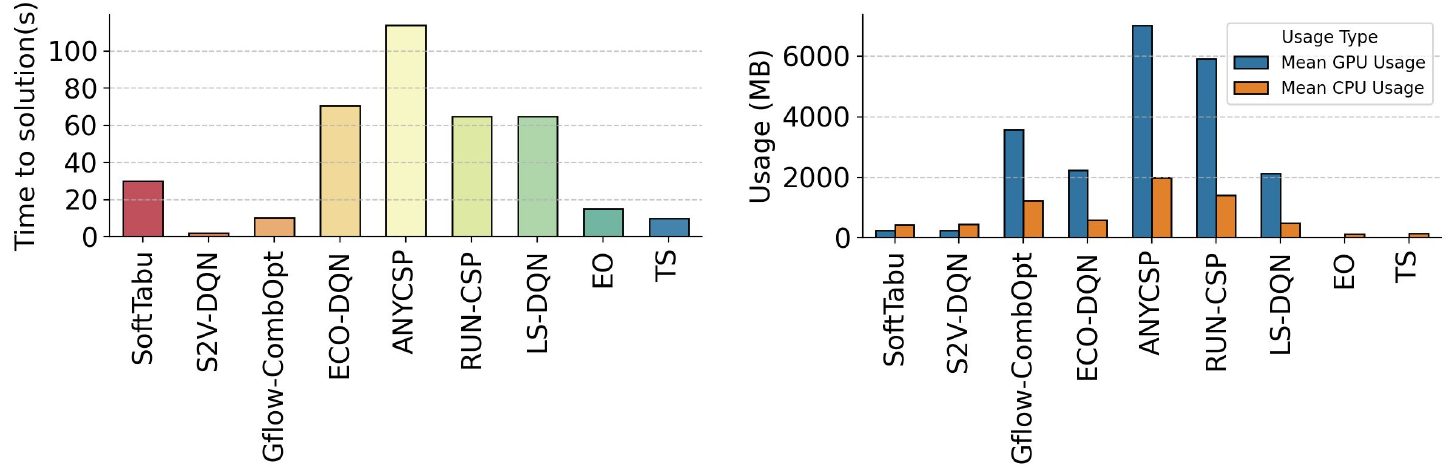}

     \caption{Comparison of the wall-clock time and average GPU and CPU memory utilization among heuristics.}
\label{fig:scalability}
\end{figure}
\textbf{Impact of Findings.} From Figure \ref{fig:scalability}, we observe that
algorithms that use multiple randomly initialized episodes require significantly more time and memory. We observe that
ANYCSP takes the longest time to complete and consumes the most memory, whereas classical heuristics are faster by an order of magnitude.



\section{Conclusion and Future Directions} \label{sec:conclusion}
In this paper, we introduce \textbf{MaxCut-Bench}, a comprehensive benchmark for evaluating deep learning-based algorithms for the Maximum Cut problem, consisting of carefully selected instance distributions and implemented algorithms.
It is our intent that \textbf{MaxCut-Bench} will foster further research and refinement of learning-based algorithms, enabling more informed evaluations and comparisons. We regard our work as a long-term evolving project and are dedicated to its continuous development. Our roadmap for the future includes expanding its scope to cover a broader spectrum of CO problems, incorporating more cutting-edge models, and integrating newer and more challenging distributions of instances.

While one might expect that extensively tailored heuristics can outperform learned approaches, our empirical
  findings suggest that simple local search heuristics frequently outperform complicated GNN-based heuristics. Specifically, Tabu Search,
  a local search heuristic that tries to avoid solutions previously encountered, outperforms all but one
  of the evaluated learned heuristics
  across a broad range of instance distributions. Even more surprising, we found that a naive reversible
  greedy algorithm RG is not consistently outperformed by several learned heuristics, including highly cited ones such as S2V-DQN. 
  In addition, we showed that ablating the GNN from ECO-DQN did not hurt its performance, and showed evidence that ECO-DQN may
  simply be learning a heuristic similar to Tabu Search. 
  On the positive side, ANYCSP did show a modest improvement over Tabu Search, although it uses many times the computational resources.
  Further, we observed that in some cases using a GNN to guide a traditional heuristic can improve the performance of the original
  heuristic.

\clearpage

\bibliographystyle{apalike} 
\bibliography{reference.bib}

\newpage


\appendix

\section{Appendix}

\subsection{Detailed Description of Datasets} \label{appendix:datasets}

In this section, we provide details on the datasets, groups of datasets, and random graphs models used in this paper.

\begin{itemize}

\item \textbf{Erd{\H{o}}s-R{\'e}nyi (ER).} This well-known random graph model by \citet{erdHos1960evolution} connects each pair of vertices with a probability \(p\). {For ease of comparison with the experiments in ECO-DQN paper, we set the same parameters from the paper} and generate graphs of size \(|V| = 200\) using \(p = 0.15\) with edge weights \(w \in \{0, \pm1\}\) for training and validation. For testing, we use $100$ test graphs from this distribution that were used for evaluating ECO-DQN. 

\item \textbf{Barab{\'a}si-Albert(BA).} The random graph model by \citet{albert2002statistical} iteratively adds nodes, connecting them to \(m\) already existing nodes. For our experiments, we generate graphs of size \(|V| = 800\) using \(m = 4\), with edge weights \(w \in \{0, \pm1\}\) and \(w \in \{0, 1\}\). For ease of comparison with the experiments in ECO-DQN paper, we also generate graphs of size \(|V| = 200\) using \(m = 4\) with edge weights \(w \in \{0, \pm1\}\) for training and validation, following \citet{barrett2020exploratory}.  For testing, we use $100$ test graphs with $200$ vertices, which were used for evaluating ECO-DQN.

\item \textbf{Holme-Kim(HK).} The random graph model by \citet{holme2002growing}, similar to the BA model, includes an extra step for each randomly created edge that forms a triangle with probability \(p\). We generate graphs of size \(|V| = 800\) using \(m = 4\) and \(p = 0.10\), with edge weights \(w \in \{0, \pm1\}\) and \(w \in \{0, 1\}\).

\item \textbf{Watts-Strogatz(WS).} The random graph model by \citet{watts1998collective} starts with a well-structured ring lattice with a mean degree of \(k\). In the next step, each edge is replaced with probability \(p\) by another edge sampled uniformly at random. This approach aims to preserve "small-world properties" while maintaining a random structure similar to ER graphs. We generate graphs of size \(|V| = 800\) using \(k = 4\) and \(p = 0.15\), with edge weights \(w \in \{0, \pm1\}\) and \(w \in \{0, 1\}\).

\item \textbf{GSet.} This dataset \citep{ye2003gset} is extensively used to benchmark {classical} heuristics \citep{benlic2013breakout, leleu2019destabilization, leleu2021scaling} for MaxCut. The dataset comprises three types of weighted and unweighted random graphs: ER graphs with uniform edge probabilities, skew graphs with decaying connectivity, and regular toroidal graphs. For generating training and validation distributions, we use the independent graph generator Rudy by Giovanni Rinaldi, which is used for generating GSet graphs as sourced from \citet{ye2003gset}. {For training and validation, we generate ER graphs of size $|V|=800$ with $p=0.06$ , union of planar graphs (skew graphs) of size $|V|=800$ with density $d=0.99$ and regular torodial  graphs of size $|V|=800$ for training and validation.}  The arguments for generating the graphs with Rudy are collected from \citet{helmberg2000spectral}.

 \item \textbf{Physics.} This distribution comes from a publicly available library of MaxCut instances\footnote{https://grafo.etsii.urjc.es/optsicom/index.php.html}
and includes synthetic and realistic
instances that are widely used in the optimization community (see references at the library website). For training and validation, we generate a similar distribution, and for testing, we
make use of a subset of the instances available, namely ten problems from Ising Spin.
glass models in physics (the first $10$ instances in Set2 of the library) following \citet{khalil2017learning}. All ten instances have $125$ nodes and $375$ edges, with edge weights \(w \in \{0,\pm1 \}\). 

\item \textbf{Sherrington-Kirkpatrick spin glass.} This distribution contains dense Sherrington-Kirkpatrick instances with elements \(J_{ij} \in \{-1,1\}\) generated from ER graphs based on examples from \citet{hamerly2019experimental}. We generate graphs of size 70 to 100 vertices for training and validation. For testing, we use instances with the best known value provided in CIM-Optimizer \citep{Chen_cim-optimizer_a_simulator_2022}.

\item  \textbf{Phase transition.} This distribution contains dense unweighted instances from ER graphs at the phase transition \((p = 0.5)\) \citep{coppersmith2004random}, based on examples from \citet{hamerly2019experimental}. We generate graphs with $100$ to $200$ vertices for training and validation and make use of test instances provided in CIM-Optimizer \citep{Chen_cim-optimizer_a_simulator_2022}.

\end{itemize}

\subsection{Detailed Description of Benchmark Algorithms} \label{appendix:algorithms}
All algorithms make use of $50$ attempts, and the best solution found is reported. The exceptions are FG and S2V-DQN, which both are deterministic and start from the empty set. Next, we provide details about each algorithm discussed in our paper.

\textbf{Traditional Heuristics.}

\begin{itemize}
    \item \textbf{Forward Greedy(FG).} The algorithm starts with an empty solution and greedily adds vertices to the solution that result in the greatest immediate increase in the objective value until no further improvements can be made. It does not allow reversible actions, meaning it does not remove any vertex from the solution once added.

    \item \textbf{Reversible Greedy(RG).} The algorithm starts with an arbitrary solution and either adds or removes a vertex, taking the action with the largest possible non-negative gain to the objective value. If all gains are negative, the algorithm terminates. 

    \item \textbf{Tabu Search(TS).} The algorithm starts with an arbitrary solution and strategically flips the membership of vertices (include/exclude from the solution set) at each step to improve the objective value. It incorporates a mechanism called tabu tenure, where, after flipping the membership of a vertex, the vertex will be marked as tabu (forbidden) for a certain period of steps to prevent the immediate reversal of the flip and encourage exploration of new areas. At each step, it flips a vertex that is not marked as tabu and results in the greatest increase in objective value. Despite being marked as tabu, a vertex can still be flipped if it leads to the best objective value discovered so far. Unlike forward and reversible greedy algorithms, this algorithm continues searching for a fixed number of iterations even when no further immediate improvements can be made, thus enhancing the search process by exploring a broader solution space and potentially finding better solutions over time. Various improved versions of this algorithm have been proposed \cite{glover1990tabu}; we consider the vanilla version of the algorithm (see Algorithm \ref{alg:tabu}). Further details can be found in \citet{glover1990tabu}.


\begin{algorithm}
\caption{Tabu Search}
\begin{algorithmic}[1] \label{alg:tabu}
\REQUIRE oracle $f$, graph $G(V,E)$, initial solution $S_0$, tabu tenure $\gamma$, maximum iterations $maxiter$
\STATE Initialize current solution $S \leftarrow S_0$
\STATE Initialize tabu list $T$ as an empty dictionary
\STATE Initialize best objective value $bestobj \leftarrow f(S)$
\STATE Initialize iteration counter $iter \leftarrow 0$
\WHILE{$iter < maxiter$}
    \STATE $bestmove \leftarrow \text{None}$
    \STATE $bestvalue \leftarrow -\infty$
    \FOR{each vertex $v \in V$}
        \STATE Flip the membership of $v$ in $S$ to obtain $S'$
        \STATE Calculate objective value $f(S')$
        \IF{$f(S') > bestobj$ \OR ($v \notin T$ \AND $f(S') > bestvalue$)}
            \STATE $bestmove \leftarrow v$
            \STATE $bestvalue \leftarrow f(S')$
        \ENDIF
    \ENDFOR
    
    \STATE Flip the membership of $bestmove$ in $S$
    \STATE Add/update $bestmove$ in tabu list $T$ with tabu tenure $\gamma$
    \IF{$f(S) > bestobj$}
        \STATE $bestobj \leftarrow f(S)$
    \ENDIF

    \FOR{each vertex $v$ in tabu list $T$}
        \STATE Decrease tabu tenure of $v$ by 1
        \IF{tabu tenure of $v$ is 0}
            \STATE Remove $v$ from $T$
        \ENDIF
    \ENDFOR
    \STATE Increment iteration counter $iter \leftarrow iter + 1$
\ENDWHILE
\RETURN $bestobj$
\end{algorithmic}
\end{algorithm}

\item \textbf{Extremal optimization (EO).} The algorithm begins with an initial arbitrary solution and sorts the vertices by their descending marginal gain. It then defines a probability distribution \(P_{k} \propto k^{-\tau}\) where \(1 \leq k \leq |V|\) for a given value of the parameter \(\tau\) to determine the likelihood of selecting each vertex based on its rank in the sorted list. At each step, an index \(k\) is selected according to this probability distribution, and the {membership of the selected vertex} is flipped. This method allows the algorithm to escape local optima and explore the search space more effectively, thereby increasing the chances of finding better solutions. Similar to TS, it stops after a fixed number of iterations (see Algorithm \ref{alg:EO}). Further details can be found in \citet{boettcher2001extremal}.

\begin{algorithm}
\caption{Extremal Optimization Algorithm}
\begin{algorithmic}[1] \label{alg:EO}
\REQUIRE oracle $f$, graph $G(V,E)$, initial solution $S_0$, tau $\tau$, maximum iterations $maxiter$
\STATE Initialize current solution $S \leftarrow S_0$
\STATE Initialize a probability distribution \(P_{k} \propto k^{-\tau}\) where \(1 \leq k \leq |V|\)
\STATE Initialize best objective value $bestobj \leftarrow f(S)$
\STATE Initialize iteration counter $iter \leftarrow 0$
\WHILE{$iter < maxiter$}

    \STATE Initialize a list $marginal\_gains \gets []$
    \FOR{each vertex $v \in V$}
        \STATE Flip the membership of $v$ in $S$ to obtain $S'$
        \STATE Calculate objective value $f(S')$
        \STATE Calculate $gain \gets f(S') - f(S)$
        \STATE Append $gain$ to $marginal\_gains$
    \ENDFOR 
    
    \STATE Sort vertices in descending order of marginal gains
    \STATE Select an index $k$ according to the probability distribution $P_k$ 
    \STATE Select the vertex $v_k$ that is in the $k$-th position in the sorted list
    \STATE Flip the membership of $v_k$ in $S$ to obtain new solution $S$
    \IF{$f(S) > bestobj$}
        \STATE $bestobj \leftarrow f(S)$
    \ENDIF

    \STATE Increment iteration counter $iter \leftarrow iter + 1$
\ENDWHILE
\RETURN $bestobj$
\end{algorithmic}
\end{algorithm}

\end{itemize}

\textbf{GNN-based heuristics}

\begin{itemize}
    \item \textbf{S2V-DQN.} Similar to FG, the algorithm starts from an empty solution and incrementally constructs solutions by {adding one vertex at each step to the current solution, guided by a GNN }. Once a vertex is added to the solution, it cannot be removed; in other words, the algorithm does not reverse its earlier decisions. The state space of its RL agent is represented by the current solution. The algorithm stops when no action can improve the objective value. The reward function of the algorithm is simply the change in the objective value. Further details can be found in \citet{khalil2017learning}.

    \item \textbf{ECO-DQN.} Unlike S2V-DQN, this algorithm starts with an arbitrary partition of vertices and allows reversible actions. \citet{barrett2020exploratory} provides seven handcrafted features per node to represent the state space of its RL agent.
At each step, it selects a vertex and flips its membership. The RL agent often chooses vertices that do not correspond to the greatest immediate increase in the objective value (non-greedy). Thus, it aims to strike a balance between exploitation and exploration of the search space. The algorithm stops after a fixed number of iterations. It provides a reward to the RL agent only when a new solution has been found, which equals the difference between the new best solution and the previous best solution. Since the reward can be very sparse, the algorithm also provides a small intermediate reward to the agent when the agent reaches a new locally minimal solution. Further details can be found in \citet{barrett2020exploratory}.

    \item \textbf{LS-DQN.}  Similar to RG, LS-DQN allows reversible actions and starts with an arbitrary solution instead of an empty one. {The state space of its RL agent is represented by the current solution. At each step, it selects a vertex and filps the membership of the selected vertex. It stops after a fixed number of iterations or can terminate on its own. The reward function of this algorithm is defined as the negative value change of the objective function at each step.} It generalizes to a variety of CO problems, like MaxCut and TSP. Further details can be found in \citet{yao2021reversible}.

    \item \textbf{Gflow-CombOpt.} The algorithm begins with an empty solution. The formulation of the Markov decision process (MDP) for the generative flow network proceeds as follows: at each step, it adds one vertex to the solution. After each action, it checks if adding the vertex would decrease the cut value. If so, it excludes the vertex, ensuring it is never added back to the solution. Despite starting with an empty solution, the algorithm generates diverse solution candidates by sampling from a probability distribution in a sequential decision-making process. Further details can be found in \citet{zhang2023let}.

    \item \textbf{RUN-CSP.} The algorithm solves CO problems that can be mapped to binary constraint satisfaction problems (CSP). It employs a graph neural network as a message-passing protocol, with the CSP instances modeled as a graph where nodes correspond to variables and edges represent constraints. Like other GNN-based heuristics, it is not a reinforcement learning approach; rather, the loss function to optimize this algorithm is designed to satisfy as many constraints as possible. The results show that it performs effectively on significantly larger instances, even when trained on relatively small ones. Further details can be found in \citet{toenshoff2021graph}.
    \item \textbf{ANYCSP.}  The algorithm is an end-to-end search heuristic for any constraint satisfaction problem. \citet{tonshoff2022one} introduced a novel representation of CSP instances, called the constraint value graph, which allows for direct processing of any CSP instance. The state space of RL agent is represented by both the current solution and the best solution found so far. At each step, the algorithm generates a soft assignment of variables within the CSP instance, enabling transitions between any two solutions in a single step. To encourage exploration and prevent the search from getting stuck in local maxima, a reward scheme similar to ECO-DQN is employed. Notably, the RL agent does not receive a reward upon reaching an unseen local minimum. Empirical evidence has shown that this approach can compete with or even surpass classical SOTA problem-specific heuristics. Further details can be found in \citet{tonshoff2022one}.
\end{itemize}

\textbf{Quantum Annealing}

Quantum annealing algorithms start by framing the optimization problem as an energy landscape of a quantum system, with the solution being the state of the lowest energy. Initially, the quantum system is set in a superposition of all possible solutions, which represents a high-energy state. The objective is to steer the system towards the lowest energy state, which corresponds to the optimal or near-optimal solution for the problem. Next, we describe two SOTA quantum annealing algorithms used in our benchmark

\begin{itemize}
    \item \textbf{Amplitude Heterogeneity Correction (AHC).}  The algorithm maps the objective function of CO problems to the energy landscape of  a physical system called Coherent Ising Machine \citep{yamamoto2017coherent}. It relaxes the binary vertex states of MaxCut problem to continuous values and find low energy states efficiently. {It finds solutions of better or equal quality for GSet instances compared to those previously known from the classical SOTA heuristic Breakout Local Search \citep{benlic2013breakout}.}
    \item \textbf{Chaotic Amplitude Control (CAC).}  {To improve the scalability, this algorithm make use of nonrelaxational dynamics that can accelerate the sampling of low energy states to reduce the time to find optimal solutions . Futher details can be found in \citet{leleu2021scaling}.}
\end{itemize}

\subsection{Baseline and Instance Bias} \label{appendix:Instance and Selection Bias}

\begin{figure}[h]
    \centering
    \includegraphics[width=\linewidth]{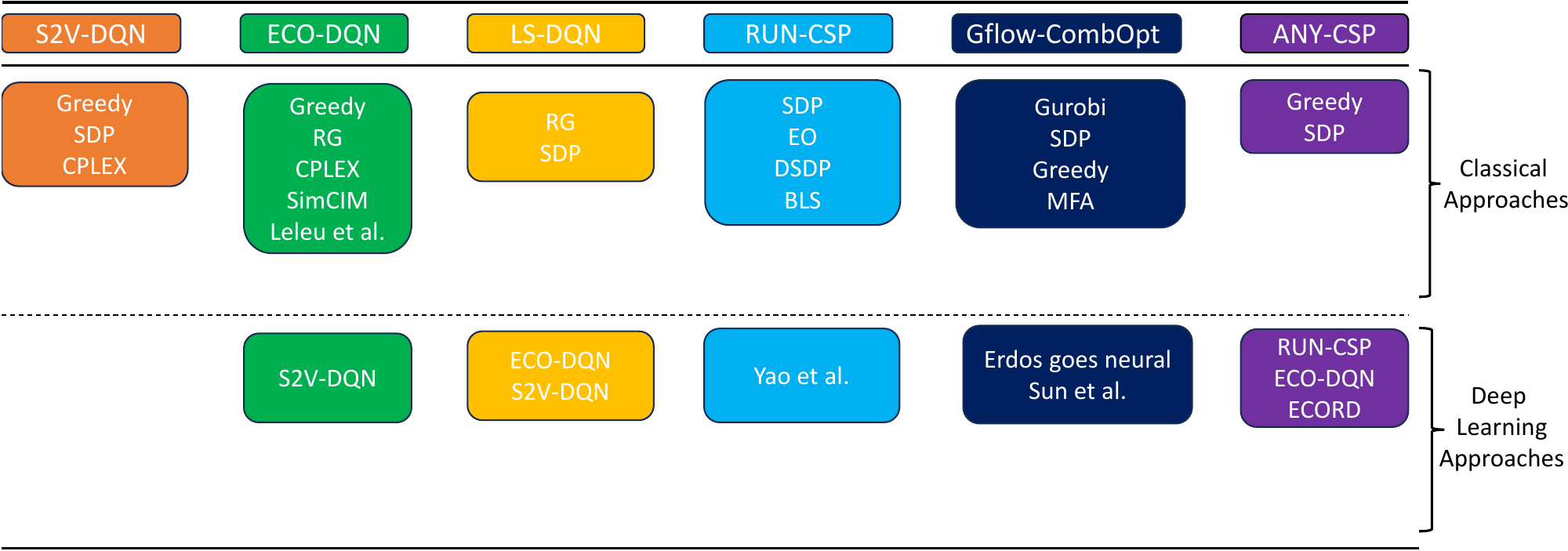}
    \caption{Researchers often select arbitrary baselines, which, when combined with instance bias, can lead to confusion in empirical evaluations.}
    \label{fig:BaselineBias}
\end{figure}

\begin{figure}[h]
    \centering
    \includegraphics[width=\linewidth]{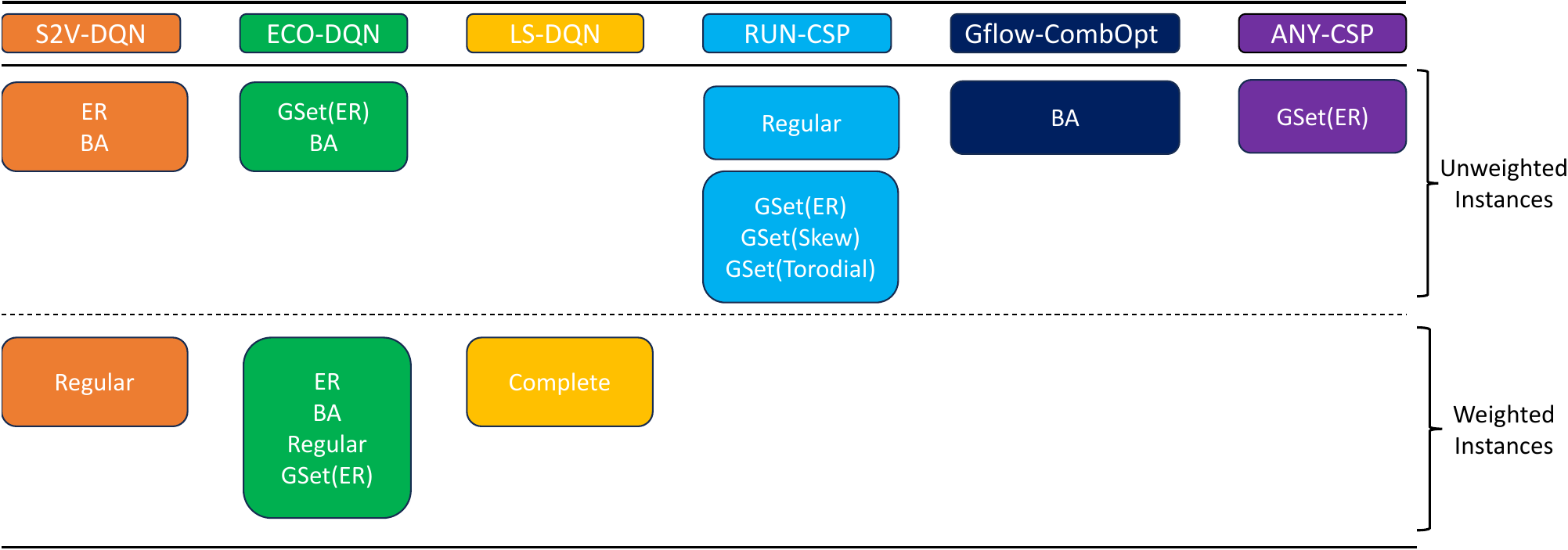}
    \caption{ Researchers evaluate their proposed learned heuristics on a diverse range of weighted and unweighted distributions for MaxCut. These instances may pose a challenge for weak baselines while potentially being relatively manageable for learned heuristics.}
    \label{fig:instance_bias}
\end{figure}

In this section, we continue our discussion on the lack of consensus regarding instances and baselines. This inconsistency leads to a situation where empirical results in different research papers are often not comparable. From Figure \ref{fig:BaselineBias}, we observe that only a handful of learned local heuristics, such as ECO-DQN and RUN-CSP, compare with SOTA classical heuristics. This highlights potential baseline biases in the current research landscape. Similarly, in Figure \ref{fig:instance_bias}, few heuristics actually evaluate their performance on hard instances that are used to benchmark SOTA heuristics. Next, we discuss the evaluation specifics for each of the learning-based algorithms.

\begin{itemize}
    \item \textbf{S2V-DQN.} As previously noted, \citet{khalil2017learning} demonstrated that a simple greedy approach, which moves a vertex from one side of the cut to the other if that action results in the greatest improvement in cut weight, is the second-best performer after S2V-DQN. Despite the well-known poor performance of this approach \citep{fujishige2005submodular} for MaxCut, we observe a lack of comparisons with simple enhancements to the greedy approach, such as iterated local search \citep{lourencco2003iterated}, TS and EO. Furthermore, other baselines like CPLEX and SDP approaches perform poorly, especially on large instances in practice, even with a longer cut-off time \citep{khalil2017learning,barrett2020exploratory}.

    \item \textbf{ECO-DQN.} Although ECO-DQN included several SOTA simulated annealing heuristics in their evaluation, they did not compare ECO-DQN with TS, the closest classical algorithm ECO-DQN resembles. Moreover, in terms of generalization, they only considered the ER distribution of the GSet. We find that the generalization performance of ECO-DQN can significantly degrade when tested on other distributions of the GSet (more details in Table \ref{tab:hard-benenchmarks}).

    \item \textbf{LS-DQN.} The performance of LS-DQN was evaluated on weighted complete graphs of various sizes, excluding standard hard instances typically used for benchmarking SOTA heuristics. This raises the possibility that on an arbitrary distribution, LS-DQN can achieve comparable performance to SOTA heuristics, while weaker baselines may struggle. Classical baselines, such as SDP and genetic algorithms, exhibited poorer performance compared to RG in the empirical evaluation of LS-DQN. However, there is a noticeable absence of comparisons with straightforward enhancements over RG. Given that LS-DQN aims to improve upon RG, it is crucial to include simple baselines that enhance the performance of RG.

    \item \textbf{RUN-CSP.} While \citet{toenshoff2021graph} included unweighted instances from the GSet distribution to evaluate GSet, our empirical evaluation indicates that both classical and learned heuristics struggle more with optimizing weighted instances compared to unweighted ones. This suggests that weighted instances pose greater optimization challenges. Evaluating the algorithm in these instances would provide a clearer understanding of its performance.  

    \item \textbf{Gflow-CombOpt.} As mentioned earlier, \citet{zhang2023let} evaluated GFlow-CombOpt for MaxCut exclusively on unweighted BA graphs, following the \citet{sun2022annealed}. These instances are neither theoretically known to be hard nor commonly used to assess the empirical performance of SOTA classical heuristics. Additionally, baselines such as Mean Field Annealing (MFA) \citep{bilbro1988optimization}, Erdos Goes Neural \citep{karalias2020erdos} and \citet{sun2022annealed}, lack empirical evidence supporting their effectiveness as heuristics for MaxCut. While Gurobi is a leading heuristic for integer programming, it often requires significant time to find an optimal solution \citep{zhang2023let}. Consequently, depending on the cutoff time applied, the best-known solution for each instance can vary considerably.

    \item  \textbf{ANYCSP.} Similar to RUN-CSP, the performance of this algorithm was evaluated on unweighted instances from the GSet distribution and compared with SOTA learned heuristics. However, classical simple heuristics were not included in the comparison, leaving it unclear whether ANYCSP can demonstrate superior performance over the simpler algorithms.
\end{itemize}


\subsection{Reproducibility} \label{appendix:reproducibility}

{In this section, we list all the changes we made to the previous implementations of algorithms and provide implementation details of the algorithms.}

\subsubsection{S2V-DQN}
We use the publicly available implementation of S2V-DQN by \citet{barrett2020exploratory} as our initial codebase\footnote{Code available at: https://github.com/tomdbar/eco-dqn} for implementing S2V-DQN. This implementation had poor scalability because it used dense representations of graphs. It also could not handle distributions containing graphs of different sizes because dense representations of graphs of different sizes cannot be batched together. Therefore, we reimplement it to ensure it can scale up easily to larger instances. To ensure a fair comparison, we evaluate the performance of our implementation against pretrained models provided by \citet{barrett2022learning} and provide the results in Table \ref{tab:eco-dqn}.

\subsubsection{ECO-DQN}

Similar to S2V-DQN, we use the publicly available implementation of ECO-DQN and make similar improvements as we did for S2V-DQN. To ensure reproducibility, we compare our implementation with the pretrained networks provided by \citet{barrett2020exploratory} and report the results in Table \ref{tab:eco-dqn}.

\subsubsection{LS-DQN}
Since the LS-DQN codebase was not publicly available initially, we contacted the authors. Unfortunately, their codebase\footnote{Code available at: https://github.com/MingzheWu418/LocalSearch-DQN} is designed specifically for clustered graphs and is not suitable for training on arbitrary graph distributions. {The authors did not provide details about } their configuration and hyper-parameter settings in the paper. However, we were able to replicate similar improvements over RG using the GNN, as reported in the paper (see Table \ref{tab:abalation}).

\subsubsection{GFlow-CombOpt}
For unweighted instances, we utilize the original implementation\footnote{Code available at: https://github.com/zdhNarsil/GFlowNet-CombOpt} of GFlow-CombOPT with the default configuration. For weighted instances, we incorporate weighted graph convolution to utilize edge weights and make necessary adjustments to ensure that a vertex is added to the solution set only if it improves the cut value compared to not including it. However, we observe poor empirical performance of the algorithm  for weighted instances. Therefore, we restrict our empirical evaluation of GFlow-CombOPT to unweighted instances.

\subsubsection{RUN-CSP and ANYCSP}

We use the PyTorch implementation of RUN-CSP\footnote{Code available at:https://github.com/toenshoff/RUNCSP-PyTorch} and ANYCSP\footnote{Code available at:https://github.com/toenshoff/ANYCSP} with the default configuration.

\subsubsection{CAC and AHC}

We use the implementations of these two algorithms provided by \citet{Chen_cim-optimizer_a_simulator_2022} and tune their hyperparameters using Bayesian Optimization Hyperband. 

\subsubsection{TS}

To {tune the value of} tabu tenure, we run a grid search with a step size of $10$, starting from $20$ to $150$, over the validation dataset and use the tuned tabu tenure for testing. We report the parameters used for our experiments in Table \ref{tab:hyparameter}.

\begin{table}[]
\centering
\caption{Parameters used for Tabu Search (TS) and Extremal Optimization (EO).}
\label{tab:hyparameter}
\begin{tabular}{@{}cccc@{}}
\toprule
Graph               & Nodes   & Tabu Tenure (TS) & Tau (EO) \\ \midrule
Gset (ER)           & 800     &                  80&          1.4\\
GSet (Skew)         & 800     &                  90&          1.4\\
BA                  & 800     &                  110&          1.3\\
WS                  & 800     &                  140&          1.4\\
HK                  & 800     &                  100&          1.4\\
Phase Transition    & 100-200 &                  20&          1.8\\ \midrule
GSet (ER)           & 800     &                  30&          1.7\\
GSet (Skew)         & 800     &                  90&          1.4\\
GSet (Torodial)     & 800     &                  100&          1.4\\
BA                  & 800     &                  120&          1.2\\
WS                  & 800     &                  110&          1.3\\
HK                  & 800     &                  110&          1.2\\
Barrett et al. (ER) & 200     &                  10&          1.9\\
Barrett et al. (BA) & 200     &                  20&          1.6\\
SK spin-glass       & 70-100  &                  20&          1.8\\
Physics (Regular)   & 125     &                  20&          1.4\\ \bottomrule
\end{tabular}
\end{table}

\subsubsection{EO}
To {tune the value} of tau, we run a grid search with a step size of $0.1$, starting from $1.1$ to $1.9$, over the validation dataset and use the tuned tau for testing. We report the parameters used for our experiments in Table \ref{tab:hyparameter}.

\subsection{Additional results on ECO-DQN}\label{appendix:eco_dqn}

The best-known solutions for the datasets used to evaluate ECO-DQN, along with the pretrained models with weights, are publicly available. We use this data to compare our results with the pretrained models and those presented in the original paper. However, {for BA60, we were unable to load the pre-trained model for S2V-DQN.} From Table \ref{tab:eco-dqn}, we observe that {our implementation} has achieved similar empirical performance. It is worth noting that for some experiments (e.g., BA200, ER100), we were unable to reproduce the results of S2V-DQN and ECO-DQN from the paper (see Table 4 in the original paper) using our implementation and pretrained models.

\begin{table}[h]
\centering
\caption{The approximation ratios, averaged across $100$ graphs for each graph structure and size.}
\label{tab:eco-dqn}
\resizebox{\textwidth}{!}{%
\begin{tabular}{@{}cccclclccccc@{}}
\toprule
      &  & \multicolumn{4}{c}{Our Implementation} &  & \multicolumn{2}{c}{Literature} &  & \multicolumn{2}{c}{Pretrained} \\ \cmidrule(lr){3-6} \cmidrule(lr){8-9} \cmidrule(l){11-12} 
Graph &  & ECO-DQN   & S2V-DQN  & SoftTabu  & TS  &  & ECO-DQN        & S2V-DQN       &  & ECO-DQN        & S2V-DQN       \\ \midrule
ER20  &  &           0.99&          0.97          &           0.99&     1.00&  & 0.99           & 0.97          &  & 0.99           & 0.97          \\
ER40  &  &           1.00&          0.97&           1.00&     1.00&  & 1.00           & 0.99          &  & 1.00           & 0.98          \\
ER60  &  &           1.00&          0.97&           1.00&     1.00&  & 1.00           & 0.99          &  & 1.00           & 0.97          \\
ER100 &  &           1.00&          0.93&           1.00&     1.00&  & 1.00           & 0.98          &  & 1.00           & 0.92          \\
ER200 &  &           1.00&          0.95&           1.00&     1.00&  & 1.00           & 0.96          &  & 1.00           & 0.95          \\ \midrule
BA20  &  &           1.00&          0.98&           1.00&     1.00&  & 1.00           & 0.97          &  & 1.00           & 0.97          \\
BA40  &  &           1.00&          0.97&           1.00&     1.00&  & 1.00           & 0.98          &  & 1.00           & 0.96          \\
BA60  &  &           1.00&          0.97&           1.00&     1.00&  & 1.00           & 0.98          &  & 1.00           & -             \\
BA100 &  &           1.00&          0.96&           1.00&     1.00&  & 1.00           & 0.97          &  & 1.00           & 0.95          \\
BA200 &  &           0.98&          0.94&           0.98&     0.98&  & 1.00           & 0.96          &  & 0.98           & 0.93          \\ \bottomrule
\end{tabular}%
}
\end{table}

Next, we revisit the generalization experiments {in the original paper} and report the results in Tables \ref{tab:ERTable} and \ref{tab:BATable}. We observe that there is no significant performance difference between TS, SoftTabu, and ECO-DQN; they all perform reasonably well. {We would also like to highlight that the generalization performance of ECO-DQN presented in the paper is somewhat inconclusive due to the lack of harder instances. Specifically, these instances may be too simple compared to the GSet instances we use for our generalization experiments, making it possible for ECO-DQN to solve them without requiring any special efforts.} However, the performance of S2V-DQN significantly drops as the size of the graphs increases. Similar results have been reported in \citet{barrett2020exploratory}.

\begin{table}[h]
\centering
\caption{Generalization of agents trained on ER graphs of size $|V |=40 $ to unseen graph sizes and structures.}
\label{tab:ERTable}
\begin{tabular}{@{}ccccc@{}}
\toprule
Graph & TS   & SoftTabu & S2V-DQN & ECO-DQN \\ \midrule
ER60  & 1.00 & 1.00     & 0.97    & 1.00    \\
ER100 & 1.00 & 1.00     & 0.96    & 1.00    \\
ER200 & 1.00 & 1.00     & 0.95    & 1.00    \\
ER500 & 0.99 & 0.99     & 0.92    & 0.99    \\ \midrule
BA40  & 1.00 & 1.00     & 0.97    & 1.00    \\
BA60  & 1.00 & 1.00     & 0.97    & 1.00    \\
BA100 & 1.00 & 1.00     & 0.94    & 1.00    \\
BA200 & 0.98 & 0.98     & 0.86    & 0.98    \\
BA500 & 0.96 & 0.98     & 0.74    & 0.97    \\ \bottomrule
\end{tabular}
\end{table}

\begin{table}[]
\centering
\caption{Generalization of agents trained on BA graphs of size $|V |=40 $ to unseen graph sizes and structures.}
\label{tab:BATable}
\begin{tabular}{@{}ccccc@{}}
\toprule
Graph & Tabu & SoftTabu & S2V-DQN & ECO-DQN \\ \midrule
ER40  & 1.00 & 1.00     & 0.97    & 1.00    \\
ER60  & 1.00 & 1.00     & 0.95    & 1.00    \\
ER100 & 1.00 & 1.00     & 0.94    & 1.00    \\
ER200 & 1.00 & 1.00     & 0.93    & 0.99    \\
ER500 & 1.00 & 0.99     & 0.90    & 0.98    \\ \midrule
BA60  & 1.00 & 1.00     & 0.96    & 1.00    \\
BA100 & 1.00 & 1.00     & 0.94    & 1.00    \\
BA200 & 0.98 & 0.98     & 0.81    & 0.98    \\
BA500 & 0.97 & 0.99     & 0.50    & 0.99    \\ \bottomrule
\end{tabular}
\end{table}

Finally, we tested agents trained on weighted ER graphs with $|V|=200$ on real-world datasets and hard instances, following the experimental setup of \citet{barrett2020exploratory}. We extended the empirical experiments of ECO-DQN by including distributions other than the ER distribution. From Table \ref{tab:hard-benenchmarks}, we observe that SoftTabu outperforms ECO-DQN, except for ER graphs where ECO-DQN performs slightly better than SoftTabu.

\begin{table}[h]
\centering
\caption{Average approximation ratios on known benchmarks: The second half of the table shows results for extended experiments.}
\label{tab:hard-benenchmarks}
\begin{tabular}{@{}ccccccc@{}}
\toprule
Dataset & Type     & Nodes & Tabu  & SoftTabu & S2V-DQN & ECO-DQN \\ \midrule
Physics & Regular  & 125  & 1.000 & 1.000    & 0.928   & 1.000   \\
G1-10   & ER       & 800  & 0.989 & 0.984    & 0.950   & 0.990   \\
G22-31  & ER       & 2000 & 0.953 & 0.977    & 0.919   & 0.981   \\ \bottomrule \midrule
G11-G13 & Torodial & 800  & 0.951 & 0.988    & 0.919   & 0.984   \\
G14-G21 & Skew     & 800  & 0.960 & 0.973    & 0.752   & 0.940   \\
G32-34  & Torodial & 2000 & 0.915 & 0.983    & 0.923   & 0.969   \\
G35-42  & Skew     & 2000 & 0.949 & 0.965    & 0.694   & 0.864   \\ \bottomrule
\end{tabular}
\end{table}
\subsection{Additional results on generalization}\label{appendix:generalization}

\begin{figure}[h]
    \centering
    \includegraphics[width=\linewidth]{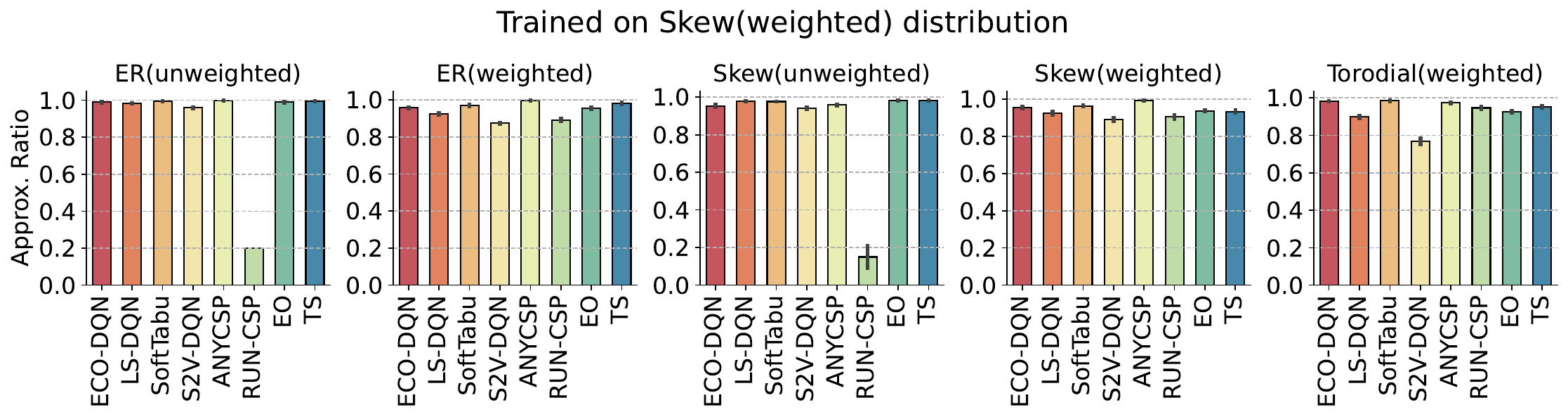}
    \caption{Generalization of agents to unseen graph sizes and structures.}
    \label{fig:appendix_skew}
\end{figure}

Due to space constraints, we present the results of the generalization of learned and classical heuristics that are trained on a skewed graph distribution of size $|V|=800$ from GSet and tested on various distributions of size $|V|=2000$ from GSet. From Figure \ref{fig:appendix_skew}, we observe that simple heuristics match or outperform the performance of learned heuristics.

\paragraph{Final Considerations.} We conclude that our implementation complies with the original implementation of ECO-DQN and conclude that ECO-DQN does not provide any significant performance boost when guided by a GNN rather than a simple linear regression.

\subsection{Additional details on Evaluation settings}\label{appendix:evaluation_settings}

Since we are using the previous codebase from several works, and they are not equally optimized, we use the number of search steps instead of a timeout. As both S2V-DQN and Gflow-CombOpt are irreversible (only add to the solution set), these algorithms can run for a maximum of $|V|$ steps. For other learned algorithms, we run the experiments for $4|V|$ steps and find no significant improvement in the objective value. We present the results of the learned heuristics in Table \ref{tab:four_times}.

\begin{table}[h]
\centering
\caption{Average approximation ratios of learned heuristics optimized for $4|V|$ steps: The first and second halves of the table show results for unweighted and weighted instances, respectively.}
\label{tab:four_times}
\begin{tabular}{@{}cccccc@{}}
\toprule
Graph               & Nodes & ECO-DQN & LS-DQN & RUN-CSP& ANYCSP \\ \midrule
Gset (ER)           &       800&         $0.997^{\pm0.001}$&        $0.993^{\pm0.001}$&        $0.979^{\pm0.002}$&        $0.999^{\pm0.000}$\\
GSet (Skew)         &       800&         $0.989^{\pm0.001}$&        $0.983^{\pm0.002}$&        $0.954^{\pm0.001}$&         $0.997^{\pm0.001}$\\
BA                  &       800&         $0.992^{\pm0.002}$&        $0.983^{\pm0.003}$&        $0.980^{\pm0.002}$&         $1.000^{\pm0.000}$\\
WS                  &       800&         $0.992^{\pm0.002}$&        $0.972^{\pm0.003}$&        $0.979^{\pm0.003}$&        $1.000^{\pm0.000}$\\
 HK& 800& $0.991^{\pm0.002}$& $0.983^{\pm0.003}$& $0.979^{\pm0.003}$&$1.000^{\pm0.000}$\\
Phase Transition    &       100-200&         $1.000^{\pm0.000}$&        $0.998^{\pm0.001}$&        $0.984^{\pm0.005}$&        $1.000^{\pm0.000}$\\  \bottomrule \midrule
GSet (ER)           &       800&         $0.981^{\pm0.006}$&        $0.950^{\pm0.008}$&        $0.912^{\pm0.009}$&        $0.998^{\pm0.002}$\\
GSet (Skew)         &       800&         $0.967^{\pm0.008}$&        $0.946^{\pm0.018}$&        $0.914^{\pm0.018}$&        $0.995^{\pm0.005}$\\
GSet (Torodial)     &       800&         $0.992^{\pm0.004}$&        $0.964^{\pm0.003}$&        $0.974^{\pm0.002}$&        $0.999^{\pm0.002}$\\
 BA& 800& $0.973^{\pm0.008}$& $0.933^{\pm0.010}$&  $0.937^{\pm0.009}$& $1.000^{\pm0.000}$\\
WS                  &       800&         $0.961^{\pm0.007}$&        $0.922^{\pm0.008}$&        $0.954^{\pm0.007}$&        $1.000^{\pm0.000}$\\
HK                  &       800&         $0.975^{\pm0.009}$&        $0.937^{\pm0.009}$&        $0.944^{\pm0.007}$&        $1.000^{\pm0.000}$\\ 
 Barrett et al. (ER)& 200& $1.000^{\pm0.001}$& $0.988^{\pm0.008}$& $0.945^{\pm0.012}$&$1.000^{\pm0.000}$\\
 Barrett et al. (BA)& 200& $0.983^{\pm0.031}$& $0.977^{\pm0.032}$& $0.960^{\pm0.016}$&$0.989^{\pm0.037}$\\
 SK spin-glass& 70-100& $1.000^{\pm0.000}$& $0.999^{\pm0.002}$& $0.962^{\pm0.019}$&$1.000^{\pm0.001}$\\
 Physics (Regular)& 125& $1.000^{\pm0.000}$& $0.995^{\pm0.009}$& $0.982^{\pm0.008}$&$1.000^{\pm0.000}$\\ \bottomrule
\end{tabular}
\end{table}

\end{document}